\documentclass[runningheads]{llncs}
\usepackage{graphicx}
\usepackage{amssymb}
\usepackage{graphicx}
\usepackage{epsfig}
\usepackage{subfigure}
\usepackage{graphicx}
\usepackage{amsmath}
\usepackage{amssymb}
\usepackage{algorithm}
\usepackage{algorithmic}
\usepackage{multirow}
\usepackage{color}

\usepackage{url}

\usepackage{array}
\newcommand{\PreserveBackslash}[1]{\let\temp=\\#1\let\\=\temp}
\newcolumntype{C}[1]{>{\PreserveBackslash\centering}p{#1}}
\newcolumntype{R}[1]{>{\PreserveBackslash\raggedleft}p{#1}}
\newcolumntype{L}[1]{>{\PreserveBackslash\raggedright}p{#1}}

\begin{document}
	

\title{Dual-Attention Graph Convolutional Network\thanks{Corresponding Author: Tong Zhang(tong.zhang@njust.edu.cn)}}


%
%
\author{Xueya Zhang\and
	Tong Zhang* \and
	Wenting Zhao \and
	Zhen Cui \and
    Jian Yang }
%

%
%


%
\authorrunning{Xueya Zhang et al.}
%
\institute{Key Lab of Intelligent Perception and Systems for High-Dimensional Information of Ministry of Education, School of Computer Science and Engineering, Nanjing University of Science and Technology\\
}

\maketitle
	
\vspace{-10pt}
\begin{abstract}
	Graph convolutional networks (GCNs) have shown the powerful ability in text structure representation and effectively facilitate the task of text classification. However, challenges still exist in adapting GCN on learning discriminative features from texts due to the main issue of graph variants incurred by the textual complexity and diversity. In this paper, we propose a dual-attention GCN to model the structural information of various texts as well as tackle the graph-invariant problem through embedding two types of attention mechanisms, i.e. the connection-attention and hop-attention,  into the classic GCN. To encode various connection patterns between neighbour words, connection-attention adaptively imposes different weights specified to neighbourhoods of each word, which captures the short-term dependencies. On the other hand, the hop-attention applies scaled coefficients to different scopes during the graph diffusion process to make the model learn more about the distribution of context, which captures long-term semantics in an adaptive way. Extensive experiments are conducted  on five widely used datasets to evaluate our dual-attention GCN, and the achieved state-of-the-art performance verifies the effectiveness of dual-attention mechanisms. 
	\keywords{dual-attention \and graph convolutional networks \and text classification.}
\end{abstract}\vspace{-8pt}

\section{Introduction}\vspace{-4pt}

Text classification is an active research field of natural language processing and multimedia, and has attracted increasing attention in recent years. For those given text sequences, the purpose of text classification is to annotate them with appropriate labels which accurately reflect the textual content. As a fundamental problem of text analysis, text classification has become an essential component in many applications, such as document organization, opinion mining, and sentiment analysis. 

To achieve classification based on texts of irregular structure, numerous algorithms have been proposed for dealing with the text classification task. Traditional methods employ hand-crafted feature, i.e. TF-IDF, bag-of-words and n-grams \cite{wang2012baselines} for text content representation, and then use widely used classifies such as support vector machine (SVM) and logistic regression (LR) for classification. However, these methods suffer from the limited feature learning ability. Deep neural network based algorithms have achieved great success in various tasks, and some studies apply them to text classification. Convolutional neural networks (CNNs) \cite{kim2014convolutional} and recurrent neural networks (RNNs) \cite{liu2016recurrent} are quite representative, which extract multi-scale features and compose them to obtain higher expressive representations. Especially, recursive RNN show better performance with the advantages of modeling sequences. However, these deep neural networks cannot well model the irregular structure of texts, which is  crucial for text recognition task. Recently, graph convolutional networks (GCNs)  \cite{bastings2017graph,kipf2016semi} have been proposed with a lot of success in various tasks, and also applied in feature representation of texts. On the other hand, due to the difficulty in modeling data variance, the attention mechanism  \cite{Mnih2014Recurrent,bahdanau2014neural,vaswani2017attention} is proposed and widely embedded in multiple models, achieving promising results on a variety of tasks.

Promising performance has been achieved on text classification by aforementioned methods, especially those GCN-based frameworks. However, challenges still exist in discriminative feature representation for describing the semantics when adapting GCN on a large number of texts. Basically, the main issue comes from the graph variants incurred by the complexity and diversity of texts, where the  variants are mainly manifested in two aspects: i)  the local connection patterns of neighbour words vary with sentences, which can not be well modeled by the uniform connection weights defined by the adjacency matrix; ii) the features of various connection scopes, which are extracted from each hop during the graph diffusion process, may contribute differently for capturing the long-term semantics in diverse texts, which make it difficult to learn the distribution of context by imposing fixed weights on them (as what is done by classic GCNs). 

In this paper, we propose a dual-attention GCN framework to deal with text classification. The proposed method can learn discriminative features from texts through inference on graphs, as well as solve the graph-invariant problem by leveraging attention mechanisms. For mining the underlying structural information of text, we construct graph models based on text sequences and further conduct graph convolution for capturing contextual information through diffusion on graphs.  Furthermore, considering the graph invariants incurred by the complexity and diversity of texts, we specifically propose two different types of attention mechanisms, i.e. the connection-attention and hop-attention, and integrate them with GCN as an whole deep framework. In view of various connection patterns between neighbour words in texts, we apply connection-attention to capture the short-term dependencies by adaptively imposing different weights specified to neighbourhoods of each word. Moreover, considering to model the long-term semantics in texts during the graph diffusion process, we propose the hop-attention which applies scaled coefficients to different scopes to make the model learn more about the distribution of context in an adaptive way. For evaluating the performance of our proposed dual-attention GCN, extensive experiments are conducted on five widely used datasets, and the experimental results  show our competitive performance comparing with those state-of-the-art methods and verify the effectiveness of the dual-attention mechanism. \vspace{-4pt}

\section{Related work}\vspace{-4pt}

Mainly two lines of research are related to our work: text classification methods from the view of application line, and graph convolution as well as attention-based methods from the view of technical line. Below we briefly overview them.

\textbf{Text classification}. Traditional methods for text classification usually concentrate on two important steps, which are split into feature engineering and classification model. For feature extraction, some hand-craft features such as TF-IDF, bag-of-words and n-grams  \cite{wang2012baselines} are very common. To classify the texts, classical machine learning methods such as logistic regression (LR) and support vector machine (SVM) did play an important role. However, the representation of text is high-demension and the neural network isn't good at processing such data, which limits the ability of feature learning. Surprisingly, deep learning methods have been proposed and successfully applied to text classification. Mikolov et.al \cite{Mikolov2013Distributed} come to focus on the model based on word embeddings and recently Shen et.al \cite{Shen2018Baseline} conduct a study between Simple Word-Embedding-based Models, which show the effectiveness of word embeddings. At the same time, the principle of some deep learning models such as CNN  \cite{lecun1998gradient} and RNN  \cite{Hochreiter1997Long}are employed to text classification. Kim et.al \cite{kim2014convolutional} led a breakthrough by directly apply CNN model to text classification. Lai et.al \cite{liu2016recurrent} successfully use a specified model LSTM to text classification, which means that CNNs and RNNs that can extract multi-scale localized spatial features and compose them to obtain higher expressive representations are suitable for the task of text classification. Effective as they are, some shortcomings are exposed immediately. They mainly capture local information so that ignore much global information such as word co-occurance.

\textbf{Graph Convolutional Network}. In recent years, graph convolutional networks (GCNs) gain more attention because of some unique advantages. Representively, Bruna et.al \cite{Bruna2014Spectral} consider possible generalizations of CNNs, which extends convolution networks to graph domains. However, the expensive computation and non-localized filter are existing problems. To address this problem, Henaff et.al \cite{Henaff2015Deep} develop an extension of Spectral Network, paying effort to spatially localizing through parameterizing spectral filters. They consider the question how to construct deep architectures with low requirements for the complexity of learning on non-Euclidean domains. On the basis of previous work, then Defferrard et.al \cite{defferrard2016convolutional} proposed a fast spectral filter, which use the Chebyshev polynomial approximation so that they are the same linear complexity of computation and  classical CNNs, and especially are suitable for any other graph structure. Subsequently, Kipf et.al \cite{kipf2016semi} change the filter to a linear function so that the performance of model won't decrease. In addition, some non-spectral methods  \cite{hamilton2017inductive,duvenaud2015convolutional} like DCNN  \cite{atwood2016diffusion} and GraphSAGE  \cite{hamilton2017inductive} make operations spatially on close neighbors. 

There are some research coming to explore the graph convolutional work that are more suitable for text classification. Firstly GCNs are used to capture the syntactic structure in  \cite{bastings2017graph}, which produce representations of words and show the improvement. The method \cite{kipf2016semi} mentioned in the last paragraph apply GCN to text classification, but it can't naturally support edge features. Some other methods like  \cite{zhang2018sentence} regard documents or sentences as the graphs of words. Differently, Yao et.al \cite{yao2018graph} propose a new way to construct the graph by regarding both documents and words as nodes, which performs quite well with GCN.

\textbf{Attention mechanism}. The attention mechanism was first proposed in the field of visual images, and  \cite{Mnih2014Recurrent} led this mechanism to become popular in the true sense. Bahdanau et.al \cite{bahdanau2014neural} use the mechanism similar to attention to simultaneously translate and align on machine translation task, which can be regarded as firstly proposing the application of the attention mechanism to NLP field. Then the Attention-based RNN model begin to be widely used in NLP, not just sequence-to-sequence models, but also for various classification problems. This mechanism can directly and flexibly capture global and local connections, and each step of the calculation does not depend on the calculation results of previous steps. Immediately, the self-attention attract people and this mechanism \cite{vaswani2017attention} also shows its effectiveness. Inspired by previous work, Veličković et.al \cite{velivckovic2017graph} propose the graph attention network applied to graph nodes with different degrees, and assign arbitrary weights that are specified to neighbors so the learning model can capture related information more precisely.

In total, we also want to learn more hidden information across edges or more effective representation of nodes, so we should consider larger scale, which means considering more contextual information. The dual-attention graph convolution network we proposed, on the one hand, the connection-attention assign different weights to nodes automatically, on the other hand, the hop-attention take some hidden information of context into account by controlling the probability of sampling pairs of nodes within some distance. 
\vspace{-4pt}

\section{The proposed method}\vspace{-4pt}\label{sec3}

\begin{figure*}[!t]
	\centering
	\includegraphics[width=0.9\linewidth]{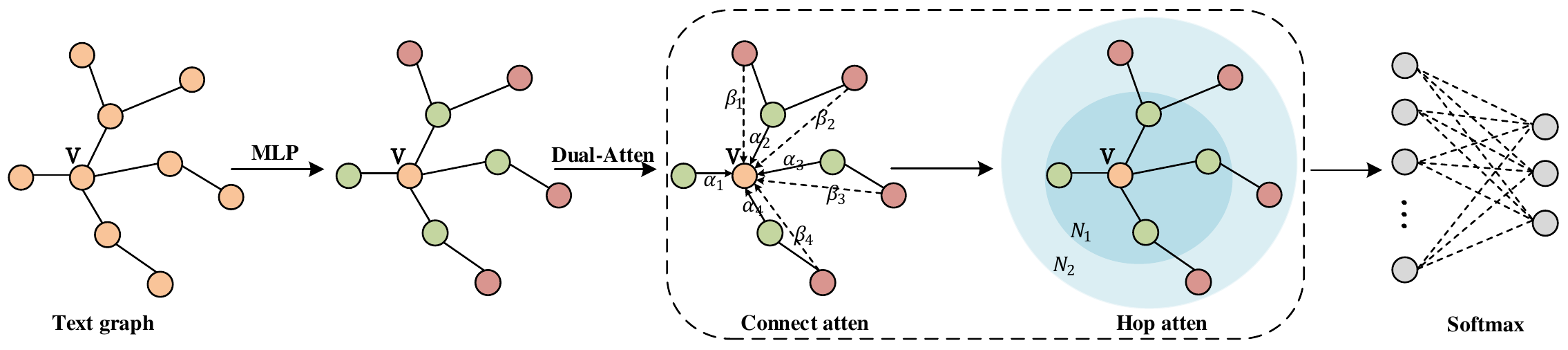}
	\caption{The working process of our dual-attention GCN. In this text subgraph, we take the node v as the center node. After a multilayer perceptron, we show the dual-attention mechanism. The connection-attention and hop-attention respectively assign weights from width and depth. $\alpha$ and $\beta$ represents the coeffients of connection-attention for different hop nodes. N represents the hop. We show the details in Section \ref{sec3}.  }
\end{figure*}\label{fig1}
In this section, we first give an overview of our proposed dual-attention GCN, and then introduce three main modules, i.e. graph construction, dual-attention layer and loss function, in detail.

\vspace{-4pt}

\subsection{Overview of dual-attention GCN}\vspace{-4pt}

The whole architecture of the proposed dual-attention GCN framework is shown in Fig.1, where the input is a graph based on a given text. For graph construction, we adopt the method proposed in  \cite{yao2018graph} which removes useless words in texts first and then models both the text and its words as nodes. This process is described in detail in Section \ref{graph_construction}.  Based on the constructed heterogeneous text graph, we conduct graph inference by passing it though our designed dual-attention GCN, where the central unit is a novel graph convolutional layer embedded with dual-attention mechanism. During the inference process, the connection-attention and hop-attention adaptively assign different weights to those neighbours of each node and the features of different scopes, respectively, to solve the graph-invariant problem (see Section \ref{Connect-attention} and Section \ref{Hop-attention} in detail). After graph inference, the obtained features are passed through a softmax classifier and finally cross entropy loss is calculated for network optimization (see Section \ref{loss}). 
\vspace{-5pt}

\subsection{Graph construction} \label{graph_construction}\vspace{-5pt}

For a given text, the corresponding graph denoted as $\mathcal{G}=(\mathcal{V},\mathcal{E}, \mathbf{A})$ is constructed for the content description, where $\mathcal{V},\mathcal{E}$ denote the sets of vertex and edges separately, and $\mathbf{A}\in{\mathbb{R}^{n\times n}}$ is the adjacency matrix describing the connection relationship between each pair of nodes. Two types of nodes are involved in $\mathcal{V}$: one type is constructed by the words in texts and the other type is constructed by the whole text itself.  To describe the relationship between these nodes,  including both the connection between the nodes of words and the connection across the nodes of a word and the text, a corpus is first built based on the training texts and the PMI and TF-IDF values are calculated based on the statistics of the corpus, which are defined as follows: 
\begin{equation*} 
\text{PMI}(i,j)=log\frac{p(i,j)}{p(i)p(j)},
\end{equation*}
and
\begin{equation*}
\text{TF-IDF}_{ij}=\text{TF}_{ij}*\text{IDF}_{i},\\
\end{equation*}
where
\begin{equation*}
\text{TF}_{ij}=n_{ij},\text{IDF}_{i}=log\frac{\mid \mathbf{D}\mid}{\{i:t_j \in \mathbf{d}_i\}}.
\end{equation*}
For PMI values, $p(i,j)=\frac{\text{W}(i,j)}{\text{W}}$ and $p(i)=\frac{\text{W}(i)}{\text{W}}$, where the statistical set W is the number of sliding windows and W(i) is the number of sliding windows that contains the word i. W(i,j) is the number of sliding windows in which word i and word j appear simultaneously. And for calculating TF-IDF values, $n_{ij}$ is the number of times the word $j$ appears in the document $i$, $\mid \mathbf{D}\mid$ is the number of documents, $\{i:t_j \in \mathbf{d}_i\}$is the the number of documents that contains the word $i$. 

Based on the defined PMI and TF-IDF values, the adjacency matrix can be obtained, which is represented as follows:
\begin{equation*}
A_{ij}=\left\{
\begin{aligned}
&\text{PMI}(i,j) && \text{$i$ and $j$ are words and PMI( $i,j$ )$>$0} \\
&\text{TF-IDF}_{i,j} && \text{$i$ is a document and $j$ is a word} \\
& 1 && i=j \\
& 0 && \text{otherwise}
\end{aligned}
\right.
\end{equation*}

\subsection{Connection-attention}\label{Connect-attention}
After obtaining the feature matrix $\mathbf{H}$=$[\vec{h}_1,...,\vec{h}_i,...,\vec{h}_j...,\vec{h}_N], \vec{h}_i\in\mathbb{R}^d$, we introduce the connection-attention mechanism to build latent representation for a specified hop . Firstly we apply a weight matrix $\mathbf{W}\in{\mathbb{R}^{d'\times d}}$ to each node, which plays a role in shared learnable linear transformation. So we obtain sufficient expressive power to transform the input features into higher-level features. We mark the new features as $\mathbf{H}'=[\vec{h}_1^{'},...,\vec{h}_i^{'},...,\vec{h}_j^{'},...,\vec{h}_N^{'}] ,  \vec{h}_i^{'}\in\mathbb{R}^{d'}$ ,where $d'$ is the new dimension of feature vectors. The connection-attention is a shared mechanism $\mathbb{R}^{d'}\times \mathbb{R}^{d'}\rightarrow\mathbb{R}$ computing the attention coefficients. The matrix $\mathbf{C}\in{\mathbb{R}^{n\times n}}$ is applied to indicate the connection-attention coefficients performing on every nodes, where $e_{ij}^{(k)}\in{\mathbf{C}}$ is the element referring to the influence of node $j$'s features to node $i$ : 
\[e_{ij}^{(k)}=a_{connection}(\vec{h}_i^{'},\vec{h}_j^{'}),{j}\in\mathcal{N}_i^k\] 
It means every node is allowed to attend the other nodes, and assigned more or less attention. Here we compute $e_{ij}^{(k)}$ for node $i$'s neighborhood in the graph, and then add a softmax function to normalize the coefficients:
\[\alpha_{ij}^{(k)}=softmax_{j}(e_{ij}^{(k)})=\frac{exp(e_{ij}^{(k)})}{\sum_{j\in\mathcal{N}_i^k}exp(e_{ij}^{(k)})}\]
In detail, attention mechanism $a_{connection}$ is expressed as:
\[\alpha_{ij}^{(k)}=\frac{exp(LeakyReLU(\vec{\mathbf{a}}^T[\vec{h}_i^{'}\rVert\vec{h}_j^{'}]))}{\sum_{j\in\mathcal{N}_i^k}exp(LeakyReLU(\vec{\mathbf{a}}^T[\vec{h}_i^{'}\rVert\vec{h}_j^{'}])}\]
where and $\rVert$ is the concatenation operation and $\vec{\mathbf{a}}$ is a weight vector. Next we apply a nonlinearity $\delta$:
\[\vec{h}_i^{'(k)}=\delta(\sum_{{j}\in\mathcal{N}_i^k}\alpha_{ij}^{(k)}\vec{h}_j^{'})\]
Now we obtain the updated feature vector $\vec{h}_i^{'(k)}$ with k-hop neighborhoods.

\subsection{Hop-attention}\label{Hop-attention}
On the basis of feature vector ${\vec{h}_i}^{(k)}$ , we consider the hop-attention by adding a constraint on the hop.  In this case, we artificially fix a coefficient set whose coeffients are according to Chebyshev inequality. We define the coeffients of hop-attention $\mathbf{Q}=(q_1,q_2,...q_c)$ , a $c$-dimensional vector regarded as the context distribution with $q_k>0$. Where $c$ is the number of hops and there is another limitation $\sum_{k}q_k=1$ . For attention mechanism $a_{hop}$ , the $\mathbf{Q}$ add weights to different range of neiborhoods, which will take more context distribution into account in a received field. For the hop-attention layer, we also add a sotfmax for regularizing :
\[\mathbf{Q}=softmax(q_1,q_2,...q_c)\] 
For $q_k\in Q$,
\[q_k=1-\frac{k-1}{c}\]
The connection-attention and hop attention work together on the nodes' features, then the feature vectors are updated : 
\[\vec{h}_i^{''}=\sum_{k=1}^cq_k\vec{h}_i^{(k)}\]
For the stability of learning process, we can adopt multi-head. In summary, we define it as:
\[\]
\[{\vec{h}_i}^{new}=\parallel_{m=1}^M\delta(\sum_{{j}\in\mathcal{N}_i^k}\alpha_{ij}^{(k,m)}\vec{h}_j^{'})\]
where $\vec{h'}_j$=$\mathbf{W}^m\vec{h}_j$ ,  $\alpha_{ij}^{(k,m)}$represents m-th dual-attention mechanism for k-hop neighbors. M is the total number of heads and $\mathbf{W}^m$ represents the corresponding weight to the m-th attention mechanism. So we can obtain a new feature vectors with the dimension of Md'.
\subsection{The loss function}\label{loss}
In this task of text classification, the documents are annoated with a single label. If the final layer, we just map $\mathbf{H}$ to the dimension of the number of classifications, then fed it into a softmax classifier.   
\[\mathbf{Z}=softmax(\mathbf{H}^{new})\]
We define the loss function by using cross-entropy as 
\[\mathcal{L}=-\sum_{d\in\mathcal{Y}_D}\sum_{f=1}^{F}Y_{df}ln\mathbf{Z}_{df}\]
where $\mathcal{Y}_D$ is the set of document indices that have labels and F is the dimension of the output features. $Y$ is the label indicator, and we add a L2 regularization.
\vspace{-5pt}

\section{Implementation Details} 
In experiments, we use pre-trained embedding features from TextGCN ~\cite{yao2018graph} with the size of 200. In the process of constructing the graph, we set the PMI window size as 20 to be more comparable. We set the learning rate as 0.05, dropout as 0.3 if not stated separately. If using the multi-head dual-attention, the dual-attention layer consists of 8 heads computing $d'$=64 dimension features and in total are 512 features. For different datasets, we fine tune the parameters. We store the graph with the form of index instead of adjacency matrix. We select fixed number nodes in specified neighborhood every time. Especially we set the batchsize as 10 and subgraph size as 200 for 20ng and MR because of the large number of nodes. The number of neighborhoods we choose 200 and learning rate set as 0.01. For the rest we select 70 nodes in the one-hop neiborhood and gather connected nodes  if we want to select two or more hop nodes. For example, the number of two-hop nodes will be the square of original data, and cube for the three-hop. Talking of the following activation we choose an exponential linear unit (ELU) ~\cite{Djork2015Fast} nonlinearity. We apply the Momentum optimizer ~\cite{Qian1999On} and models are trained to minimize cross-entropy with 300 epochs.

\section{Experiments}

\vspace{-5pt}
\subsection{Datasets}
We also ran our experiment on the five used benchmark corpora, including 20-Newsgroups, Ohsumed, R52, R8 and Movie Review(MR). R52 and R8 are two subsets of the Reuters 21578 dataset. The datasets processed are same as  \cite{yao2018graph}, and we summarize the interesting characteristics of them in Table 1. 

The 20NG consists of 18846 documents from 20 different newsgroups. In this dataset, training set includes 11314 documents and test set includes 7532 documents. The Ohsumed is a bibliographic dataset of medical literature. We just focus on the single-labeled documents from 23 disease categories. There are 3357 documents in the training set and 4043 documents in the test set. R52 and R8 are selected from the Reuters 21578 dataset. They have 52 and 8 categories respectively. R52 is divided  documents to training set and  documents to test set. R8 has the training set of 5485 documents and the test set of 2189 documents. For MR, it's a movie review dataset that only contains two classification. The MR is split to 7108 training documents and 3554 test documents. All the datasets were processed by cleaning the text, where stop words defined in NLTK were removed. Additionally, the words appear less than five times for 20NG, Ohsumed, R52 and R8 are also taken away except MR. Because of the short document, we keep the words appearing less than 5 times. As shown in Table 1,we summarize the division of each data set.
\begin{table}
	\caption{Details of datasets}
	\label{tab:data}
	\centering
	\begin{tabular}{C{1.5cm}C{1.5cm}C{1.5cm}C{1.5cm}C{1.5cm}C{1.5cm}}
		\hline
		Dataset & train  &words& test & nodes & classes \\
		\hline
		20ng  & 11314 &42757& 7532 & 61603  &  20\\
		mr  & 7108 &18764& 3554 & 29426  &   2 \\
		ohsumed & 3357 &14157& 4043 & 21557  &  23  \\
		R52 & 6532 &8892& 2568 & 17992  &  52 \\
		R8 & 5485 &7688& 2189 & 15362 &   8  \\
		\hline
	\end{tabular}
\end{table}

In experiments, the method is applied to the five datasets to complete the task of text classification. Additionally, we explore the effectiveness of our dual-attention GCN by comparing the results with ourselves, and experiment on the hop K to determine what value is appropriate.
\subsection{Results and comparisons}\vspace{-5pt}

\begin{table}
	\caption{Performance(\%) on five datasets: 20NG, MR, Ohsumed, R52 and R8. "--" donates the original paper didn't report the results. }
	\label{tab:performance}
	\centering
	\begin{tabular}{C{4cm}C{1.5cm}C{1.5cm}C{1.5cm}C{1.5cm}C{1.5cm}}
		\hline
		METHODS & 20NG & MR & Ohsumed & R52 & R8\\
		\hline
		TF-IDF+LR & 83.19 & 74.59 & 54.66 & 86.95 & 93.74   \\
		CNN-rand & 76.93 & 74.98& 43.87& 85.37& 94.02   \\
		CNN-non-static & 82.15 & \textbf{77.75}& 58.44& 87.59& 95.71   \\
		LSTM & 65.71 & 75.06& 41.13& 85.54 & 93.68  \\
		LSTM(pre-trained) &75.43 &77.33&51.10&90.48&96.09  \\
		PV-DBOW & 74.36 &61.09&46.65&78.29&85.87  \\
		PV-DM&51.14&59.47&29.50&44.92&52.07\\
		PTE&76.74&70.23&53.58&90.71&96.69 \\
		fastText&79.38&75.14&57.70&92.81&96.13\\
		fastText(bigrams)&79.67&76.24&55.69&90.99&94.74\\
		SWEM&85.16&76.65&63.12&92.94&95.32\\
		LEAM&81.91&76.95&58.58&91.84&93.31\\
		Graph-CNN-C&81.42&77.22&63.86&92.75&96.99\\	
		Graph-CNN-S&--&76.99&62.82&92.74&96.80\\	
		Graph-CNN-F&--&76.74&63.04&93.20&96.89\\
		TextGCN&86.34&76.74&68.36&93.56&97.07\\
		\hline
		OURS(dual-attention)&\textbf{87.00}&77.14&\textbf{69.19}&\textbf{93.58}&\textbf{97.36}\\	
		\hline
	\end{tabular}
\end{table}
We compare our proposed method dual-attention GCN with multiple state-of-the-art text classification and embedding methods by following   , including TF-IDF+LR \cite{yao2018graph}, CNN  \cite{kim2014convolutional} , LSTM  \cite{liu2016recurrent} , Bi-LSTM, PV-DBOW  \cite{Le2014Distributed} , PV-DM  \cite{Le2014Distributed} , PTE  \cite{tang2015pte} , fastText  \cite{joulin2016bag} , SVEM \cite{Shen2018Baseline} , LEAM  \cite{peng2018large} , Graph-CNN-C  \cite{defferrard2016convolutional}, Graph-CNN-S  \cite{Bruna2014Spectral} , Graph-CNN-F  \cite{Henaff2015Deep} and TextGCN. TF-IDF+LR is the bag-of-words model set term frequency-inverse document frequency as weights with Logistic Regression classifier. CNN is the Convolutional Neural Network and in experiment and explored with CNN-rand and CNN-non-static. CNN-rand uses the word embeddings initialized randomly and CNN-non-static uses the word embeddings pre-trained. The word embeddings of LSTM is processed the same as CNN. Bi-LSTM is a bi-directional LSTM using pre-trained word embeddings. PV-DBOW and PV-DM are paragraph vector models and followed the Logisitic Regression classifier. The obvious difference is that the former considers the orders of the words but the latter does not. PTE is predictive text embedding, which using the graph included word, documents and labels and later regarding the average of word embeddings as document embeddings. The fastText also use the average of word or n-grams embeddings to generalize document embeddings, and in experiment we try the bigrams and non-bigrams. SWEM is a word embedding model and LEAM is a label-embedding attentive model. For Graph-CNN-C, Graph-CNN-S and Graph-CNN-F, they are all graph CNN models that operate on graphs with word embeddings. The difference from the three is that they use different filters, respectively, Chebyshev filter, Spline filter and Fourier filter. TextGCN aims to construct a heterogeneous text graph containing words and documents. For our dual-attention GCN, we use embedding features and run 10 times. We show the mean of 10 results, and especially compare the results of our model without dual-attention mechanism.

The details of comparison results are reported in Table 2. We show observations as follows:
\begin{itemize}
	\item {} The TF-IDF+LR shows good performance and especially on the 20NG. It even performs better than some deep learning models. The simple method that increases words' importance with the number of times they appear in the file seems to be more suitable for the long texts. But not reflecting the position information of the word also limits the continued growth of accuracy. 
	\item {} For the CNN and LSTM, it's obvious that two models were enhanced by using pre-trained word embedding features. CNN with randomly initialized embeddings and LSTM using the last hidden state as the representation of text perform not as well as using pre-trained word embeddings. One thing they have in common is that they perform better than TF-IDF+LR on short texts but worse than long texts. 
	\item {} Conversely, the performance of PV-DBOW seems to be better on the long texts like 20NG and PV-DM seems terrible. It's likely to be the reason that PV-DBOW sampled words randomly from the output paragraph and ignoring the word orders. But PV-DM shows effect on MR with taking word orders into account and exactly the word orders are more necessary to focus on.
	\item {} The performance of PTE and fastText are more satisfied, which might because PTE is a semi-supervised representation learning method for text data and fastText is supervised. However the CNN with pre-trained embeddings still outperforms and might because CNN model can handle labeled information more effectively by utilizing word orders in the local context and solve the ambiguity of the word sense.
	\item {} There is a significant improvement in SWEM and LEAM, the simple word-embedding based model and the the joint embedding of words and labels model, which indicate the pooling operations and considering nonlinear interaction between phrase and labels do play a role.
	\item {} The graph CNN model with three kinds of filters show more competitive results on the five datasets. The results demonstrate that these supervised models are really suitable for the graph or node-focused applications. Except the long text 20NG, the overall perform is very well on the other four datasets.
	\item {} In contrast, our proposed method is superior to multiple state-of-the-art on the datasets 20NG, Ohsumed, R52 and R8. At the same time, it's also show the competitive performance on the dataset MR. Especially, the results are improved significantly on 20NG and Ohsumed. With the embedding features including the relations of word-word and document-word, our dual-attention mechanism can not only dynamically assign weights to related nodes and learn the edges, but also emphasize the context distribution. So it outperforms shown methods on both long texts and short texts. It's because that it equals to selecting the nodes that are more worthy of attention and update self by using their features. 	
\end{itemize}

\textbf{Comparisons on the hop K}

For the hop-attention K, determining the hop K is equivalent to determining the size of the receptive field, that is, the length of the context. So the most appropriate size of K should be a problem that needs to be concerned. We experiment on the dataset Ohsumed and R52 to explore the impact of K on classification. To make the difference more obvious, we set one-hop neighbor nodes to 10. 

As shown in Table  \ref{tab:comparek}, we can find that on Ohsumed the model performs best when K is 3, which means context distribution is ignorable. The performance can be improved about 3\% between the hop of one and two. But for the hop three, the difference isn't obvious as before. For example, in the sentence "I am in my study, surrounded by books.", we need the directly adjacent word "my" to confirm "study" is a noun but we can't know it means learning or the room. "Surrounded by books" helps to understand the real meaning of "study" and these words are enough to help understand the meaning of word "study". For the shorter text R52, we can see that the model shows best performance when the hop K is 2. According to this result, we can note that two hops are more suitable for R52 and it might because the text is shorter than Ohsumed. When capturing information from other nodes, excess information will disturb the classification. The results above demonstrate the effectiveness of our hop-attention, which concerns nodes differently at different distances.
\begin{table}
\caption{Comparisions on the hop K on datasets Ohsumed and R52}
\label{tab:comparek}
\centering
\begin{tabular}{C{2cm}C{2cm}C{2cm}}
	
	\hline
	K & Ohsumed(\%) & R52(\%) \\
	\hline
	1  & 52.44 &89.71 \\
	2  & 55.16&\textbf{90.34} \\
	3 & \textbf{55.47}&88.94  \\
	\hline
\end{tabular}
\end{table}
\subsection{Ablation Study}
To better show the effectiveness of our models, we continue to do a comparison experiment, that is, keep the other settings unchanged and only remove the dual-attention mechanism on the five data sets to compare. We first conduct the experiment with a convolution layer by commenting out the dual-attention section. Then we recovery our dual-attention mechanism to compare. As shown in Table  \ref{tab:ablation}, the performance is crucial for the model without dual-attention mechanism. Although the dataset R8 has reached a high accuracy rate, our method still pulls it up a bit. In total, the performance can be improved by about 1\% and 5\% for the task of text classification.  

In contrast, there is a significant difference between adding and not adding dual-attention mechanism to long texts or short texts, which demonstrates that our dual-attention GCN can capture both short-term dependences and long-term dependences well. It indicates that our dual-attention GCN model the structural information of various texts well. The connection-attention adaptively assign weights to related words and the hop-attention learn more about the distribution of context, so whether the long texts or short texts are both classified well.
\begin{table*}[htb!]
	\caption{Self comparison on 20NG, MR, Ohsumed, R52 and R8 }
	\label{tab:ablation}
	\centering
	\begin{tabular}{C{2cm}C{2cm}C{2cm}}
		
		\hline
		Dataset &  dual-attention  & convolution \\
		\hline
		20NG  & \textbf{87.00} & 84.46 \\
		MR  & \textbf{77.14} & 74.25 \\
		Ohsumed & \textbf{69.19} & 64.95  \\
		R52 & \textbf{93.58}& 89.17  \\
		R8 & \textbf{97.36} & 96.71   \\
		\hline
	\end{tabular}
\end{table*}

\section{Conclusion} \vspace{-4pt}
In this paper, we propose a dual-attention graph convolutional network and apply it to text classification. We aim to encode various connection patterns between related nodes and learn more about the context distribution. To this end, we adopt the connection-attention and the hop-attention, one dynamically assigns weights, and one considers the importance of context. Our dual-attention graph convolutional network model the structural information of various texts and adaptively learn the representation of text. In experiment, we verified the effectiveness of our model on five widely used dataset, and further compare with the model removing the dual-attention. The comparison also shows that our model is very powerful in contrast with the non-dual-attention model and is very suitable for both long and short texts. In total, they are effective and can achieve the performance of state-of-the-art.
\section*{Acknowledgements}
This work was supported by the National Natural Science Foundation of China (Grants Nos.  61772276, 61972204, 61906094), the Natural Science Foundation of Jiangsu Province (Grant No. BK20190452), the fundamental research funds for the central universities (No. 30919011232).

\bibliographystyle{splncs04}
\bibliography{sample-base.bib}

\begin{thebibliography}{10}
\providecommand{\url}[1]{\texttt{#1}}
\providecommand{\urlprefix}{URL }
\providecommand{\doi}[1]{https://doi.org/#1}

\bibitem{atwood2016diffusion}
Atwood, J., Towsley, D.: Diffusion-convolutional neural networks. In: Advances
  in Neural Information Processing Systems. pp. 1993--2001 (2016)

\bibitem{bahdanau2014neural}
Bahdanau, D., Cho, K., Bengio, Y.: Neural machine translation by jointly
  learning to align and translate. ICLR  (2015)

\bibitem{bastings2017graph}
Bastings, J., Titov, I., Aziz, W., Marcheggiani, D., Sima'an, K.: Graph
  convolutional encoders for syntax-aware neural machine translation. arXiv
  preprint arXiv:1704.04675  (2017)

\bibitem{Bruna2014Spectral}
Bruna, J., Zaremba, W., Szlam, A., Lecun, Y.: Spectral networks and locally
  connected networks on graphs. Computer Science  (2014)

\bibitem{Djork2015Fast}
Clevert, D.A., Unterthiner, T., Hochreiter, S.: Fast and accurate deep network
  learning by exponential linear units (elus). Computer Science  (2015)

\bibitem{defferrard2016convolutional}
Defferrard, M., Bresson, X., Vandergheynst, P.: Convolutional neural networks
  on graphs with fast localized spectral filtering. In: Advances in neural
  information processing systems. pp. 3844--3852 (2016)

\bibitem{duvenaud2015convolutional}
Duvenaud, D.K., Maclaurin, D., Iparraguirre, J., Bombarell, R., Hirzel, T.,
  Aspuru-Guzik, A., Adams, R.P.: Convolutional networks on graphs for learning
  molecular fingerprints. In: Advances in neural information processing
  systems. pp. 2224--2232 (2015)

\bibitem{hamilton2017inductive}
Hamilton, W., Ying, Z., Leskovec, J.: Inductive representation learning on
  large graphs. In: Advances in Neural Information Processing Systems. pp.
  1024--1034 (2017)

\bibitem{Henaff2015Deep}
Henaff, M., Bruna, J., Lecun, Y.: Deep convolutional networks on
  graph-structured data. Computer Science  (2015)

\bibitem{Hochreiter1997Long}
Hochreiter, S., Schmidhuber, J.: Long short-term memory. Neural Computation
  \textbf{9}(8),  1735--1780 (1997)

\bibitem{joulin2016bag}
Joulin, A., Grave, E., Bojanowski, P., Mikolov, T.: Bag of tricks for efficient
  text classification. arXiv preprint arXiv:1607.01759  (2016)

\bibitem{kim2014convolutional}
Kim, Y.: Convolutional neural networks for sentence classification. arXiv
  preprint arXiv:1408.5882  (2014)

\bibitem{kipf2016semi}
Kipf, T.N., Welling, M.: Semi-supervised classification with graph
  convolutional networks. arXiv preprint arXiv:1609.02907  (2016)

\bibitem{Le2014Distributed}
Le, Q., Mikolov, T.: Distributed representations of sentences and documents.
  In: International conference on machine learning. pp. 1188--1196 (2014)

\bibitem{lecun1998gradient}
LeCun, Y., Bottou, L., Bengio, Y., Haffner, P., et~al.: Gradient-based learning
  applied to document recognition. Proceedings of the IEEE  \textbf{86}(11),
  2278--2324 (1998)

\bibitem{liu2016recurrent}
Liu, P., Qiu, X., Huang, X.: Recurrent neural network for text classification
  with multi-task learning. arXiv preprint arXiv:1605.05101  (2016)

\bibitem{Mikolov2013Distributed}
Mikolov, T., Sutskever, I., Chen, K., Corrado, G., Dean, J.: Distributed
  representations of words and phrases and their compositionality. Advances in
  Neural Information Processing Systems  \textbf{26},  3111--3119 (2013)

\bibitem{Mnih2014Recurrent}
Mnih, V., Heess, N., Graves, A., et~al.: Recurrent models of visual attention.
  In: Advances in neural information processing systems. pp. 2204--2212 (2014)

\bibitem{peng2018large}
Peng, H., Li, J., He, Y., Liu, Y., Bao, M., Wang, L., Song, Y., Yang, Q.:
  Large-scale hierarchical text classification with recursively regularized
  deep graph-cnn. In: Proceedings of the 2018 World Wide Web Conference on
  World Wide Web. pp. 1063--1072. International World Wide Web Conferences
  Steering Committee (2018)

\bibitem{Qian1999On}
Qian, N.: On the momentum term in gradient descent learning algorithms. Neural
  Netw  \textbf{12}(1),  145--151 (1999)

\bibitem{Shen2018Baseline}
Shen, D., Wang, G., Wang, W., Min, M.R., Su, Q., Zhang, Y., Li, C., Henao, R.,
  Carin, L.: Baseline needs more love: On simple word-embedding-based models
  and associated pooling mechanisms. arXiv preprint arXiv:1805.09843  (2018)

\bibitem{tang2015pte}
Tang, J., Qu, M., Mei, Q.: Pte: Predictive text embedding through large-scale
  heterogeneous text networks. In: Proceedings of the 21th ACM SIGKDD
  International Conference on Knowledge Discovery and Data Mining. pp.
  1165--1174. ACM (2015)

\bibitem{vaswani2017attention}
Vaswani, A., Shazeer, N., Parmar, N., Uszkoreit, J., Jones, L., Gomez, A.N.,
  Kaiser, {\L}., Polosukhin, I.: Attention is all you need. In: Advances in
  neural information processing systems. pp. 5998--6008 (2017)

\bibitem{velivckovic2017graph}
Veli{\v{c}}kovi{\'c}, P., Cucurull, G., Casanova, A., Romero, A., Lio, P.,
  Bengio, Y.: Graph attention networks. arXiv preprint arXiv:1710.10903  (2017)

\bibitem{wang2012baselines}
Wang, S., Manning, C.D.: Baselines and bigrams: Simple, good sentiment and
  topic classification. In: Proceedings of the 50th annual meeting of the
  association for computational linguistics: Short papers-volume 2. pp. 90--94.
  Association for Computational Linguistics (2012)

\bibitem{yao2018graph}
Yao, L., Mao, C., Luo, Y.: Graph convolutional networks for text
  classification. arXiv preprint arXiv:1809.05679  (2018)

\bibitem{zhang2018sentence}
Zhang, Y., Liu, Q., Song, L.: Sentence-state lstm for text representation.
  arXiv preprint arXiv:1805.02474  (2018)

\end{thebibliography}

\end{document}